\documentclass{article}
\usepackage{spconf,amsmath,graphicx}

\usepackage{multirow}
\usepackage{array,booktabs,arydshln}

\usepackage{caption}
\usepackage{subcaption}
\usepackage{pgfplots}
\usepackage{standalone}

\usepackage{cite}

\usepackage{subcaption}
\usepackage{caption}

\usepackage{algorithm}
\usepackage{algpseudocode}
\usepackage{bbm}
\usepackage{graphicx}
\usepackage{hyperref}
\usepackage[colorinlistoftodos,prependcaption,textsize=tiny]{todonotes}
\usepackage{amsfonts}
\usepackage{amssymb}
\usepackage{amsbsy}
\usepackage{amsthm, mathrsfs}
\usepackage{mathtools}

\usepackage{longtable}
\algnewcommand\algorithmicinput{\textbf{Input:}}
\algnewcommand\INPUT{\item[\algorithmicinput]}
\algnewcommand\algorithmicoutput{\textbf{Output:}}
\algnewcommand\OUTPUT{\item[\algorithmicoutput]}
\title{Reconstruction of privacy-sensitive data from protected templates}
\name{ Shideh Rezaeifar, Behrooz  Razeghi, Olga Taran, Taras Holotyak, Slava Voloshynovskiy\thanks{S. Rezaeifar and O. Taran are  supported by the SNF project No 200021\_182063, B.~Razeghi is by the ERA-Net project ID\_IoT No 20CH21\_167534. S. Voloshynovskiy is a corresponding author. 
}}
\address{University of Geneva\\
$\lbrace$shideh.rezaeifar, behrooz.razeghi, olga.taran, taras.holotyak, svolos$\rbrace$@unige.ch
}
\begin{document}
%\ninept
%
\maketitle
\begin{abstract}
 
In this paper, we address the problem of data reconstruction from privacy-protected templates, based on recent concept of sparse ternary coding with ambiguization (STCA). The STCA is a generalization of randomization techniques which includes random projections, lossy quantization, and addition of ambiguization noise to satisfy the privacy-utility trade-off requirements. The theoretical privacy-preserving properties of STCA have been validated on synthetic data. However, the applicability of STCA to real data and potential threats linked to reconstruction based on recent deep reconstruction algorithms are still open problems. Our results demonstrate that STCA still achieves the claimed theoretical performance when facing deep reconstruction attacks for the synthetic i.i.d. data, while for real images special measures are required to guarantee proper protection of the templates.

\end{abstract}
\begin{keywords}
Privacy, template protection, reconstruction, ambiguization, deep learning.
\end{keywords}

%%%%%%%%%%%
%

\vspace{-4pt}

\section{Introduction}
\label{sec:intro}

\vspace{-6pt}

Modern machine learning is based on the usage of massive data sets that often contain privacy-sensitive  information. A similar problem exists with biometrics that are used in both private security systems granting access to various devices and services, and public security systems covering various surveillance and monitoring applications. In recent years, the advancement of personalized medicine applications also requires reliable privacy protection of genetic data and privacy sensitive clinical records. Despite the broad variety of these applications, machine learning tools are more often used to extract templates from the data that by itself does not guarantee their privacy protection against reconstruction attacks. At the same time, it is demonstrated that the original data can be reliably reconstructed from templates and non-linear representations extracted by both hand-crafted methods based on local descriptors \cite{dosovitskiy2016inverting} and deep representations \cite{mai2018reconstruction}. Once successfully reconstructed, an adversary might use these data to impair both privacy and security in the above applications.

The problem of template privacy protection in view of reconstruction attacks is well recognized and various generic methods were proposed such as fuzzy commitment schemes \cite{dodis2004fuzzy} and secure sketches \cite{bringer2008theoretical}, helper data based methods \cite{verbitskiy2003reliable,ignatenko2009biometric}, concealable template protection and robust hashing \cite{sutcu2005secure} as well as several practical methods in biometrics applications \cite{mai2018reconstruction}. We do not pretend to be exhaustive in our overview and refer interesting readers to \cite{nandakumar2015biometric}. 
Recently, a concept of the STCA was proposed that combines and extends the encoding and randomization principles from the information-theoretic perspectives \cite{Razeghi2017wifs, Razeghi2018icassp, Razeghi2018eusipco}. The STCA ensures the protection of both templates and queries in authentication and identification systems against the adversarial reconstruction and clustering \cite{Razeghi2017wifs, Razeghi2018icassp}. 
%Moreover, it avoids adversarial clustering of protected data sets. 
In contrast to binary templates, the STCA is based on sparse ternary encoding \cite{Ferd1706:Sparse} of data ensuring the maximum information preservation for the authorized users, while minimizing the leakages for unauthorized users using a special ambiguization scheme based on an addition of noise to the zero components of ternary codes that do not contain any significant information. The authorized users benefit from the presence of degraded authentic data in a form of probe allowing the reliable reconstruction of the original features with a minimum loss of information. Thus, the STCA enables the verification and identification in the original space in contrast to the binarized templates, which face the loss of information, and yet enjoying fast search and optimal performance \cite{Razeghi2017wifs, Razeghi2018icassp, Razeghi2018eusipco}. 

\begin{figure*}
    \centering
    \includegraphics[scale=0.37]{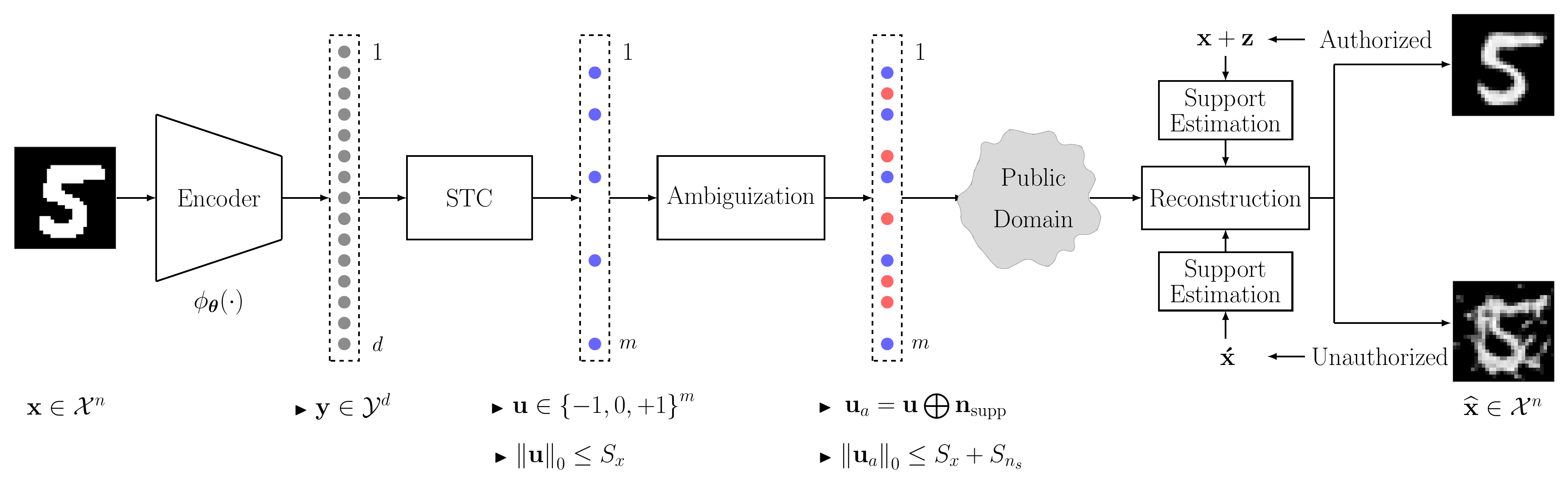}
    \vspace{-9pt}
    \caption{General block diagram explaining the adversarial reconstruction of $\bf \hat x$ based on publicly available privacy protected template ${\bf u}_a $ based on the STCA applied to the template $\bf y$ extracted from $\bf x$ using an extractor $\phi_{\boldsymbol \theta}(\cdot)$.}
    \vspace{-15pt}
    \label{Fig:GeneralBlockDiagram}
\end{figure*}

\vspace{-3pt}

To the best of our knowledge, the STCA was not investigated under the deep adversarial reconstruction. Furthermore, the methods of template reconstruction based on the deep machine learning techniques considered in \cite{dosovitskiy2016inverting} and \cite{mai2018reconstruction} were not investigated under the advanced privacy-preserving methods. Therefore, the goal of this paper is to make one step forward and consider the generalized reconstruction from the STCA protected data under adversarial deep reconstruction attacks. Along this way, we target to practically confirm the achievability of theoretical limits based on previously reported results \cite{Razeghi2017wifs, Razeghi2018icassp}. For this purpose, we will use synthetic data and analytically treatable feature extraction methods from one side and real data from another one. The adversarial attacks will be investigated in two settings of reconstruction from the protected template with all `known model parameters' besides the randomization noise, that will be kept secret for the attacker, and `unknown model parameters'. 
%\textcolor{red}{We will also attempt at demonstrating various possibilities of reconstruction facing technical difficulties related to the problematic differentiation of quantization operators and lack of image priors.} \textcolor{blue}{Olga, We did?}

%\vspace{-1pt}
\textit{Notations:} $X$ and $\mathbf{X}$ denote random variables and random vectors, while their realizations are denoted as $x$ and $\mathbf{x}$, respectively. %All matrices use capital and up-right font as in $\mathrm{X}$.

%\vspace{-1pt}
\textit{Paper Organization:} 
%The rest of paper is organized as follows.
The problem formulation is given in Section \ref{sec:problem}. The reconstruction under the known model parameters is addressed in Section \ref{sec:recovery_known} and under the unknown parameters in Section \ref{sec:recovery_unknown}. Section \ref{sec:Experimental-results} presents the experimental results and Section \ref{Conclusions} concludes the paper.

%VGG model was known
%
%no privacy-protection; randomization is mentioned as future research
%
%STC: protection with randomization: random projections, ternary signals and ambiguization
%
%Recently proposed privacy privacy protection mechanism. The STC has several remarkable properties that it preserves information for the authorized users while minimizing for the unauthorized ones.

\vspace{-7pt}

\section{Problem formulation}
\label{sec:problem}

\vspace{-6pt}

Consider a signal ${\bf x} \in {\mathcal{X}}^n$, where $\mathcal{X} \subset \mathcal{R}$. %%% Template extraction %%%% 
A generic template extraction scheme is denoted as:\vspace{-6pt}
\begin{equation}
\label{eq:template extraction}
{\bf y} = \phi_{\boldsymbol \theta}({\bf x}),
\end{equation}
where $ \phi_{\boldsymbol \theta}({\bf x}) \! \! = \!\! \sigma_L(W_L...\sigma_1(W_1{\bf x}))$ denotes any function in the form of a deep network parametrized by  some learnable parameters ${\boldsymbol \theta} \! = \!\!  \{W_1,\cdots \!,W_L\}$ with a set of element-wise non-linearities $\sigma_l(\cdot)$, $1 \! \leq  \! l \! \leq  \! L$, such as VGG$\_$19 \cite{simonyan2014very}, AlexNet \cite{krizhevsky2012imagenet}, etc. Furthermore, the template ${\bf y}$ can also be raw data, extracted features using any known hand crafted methods, aggregated local descriptors based on BoW, FV, VLAD \cite{jegou2009burstiness, perronnin2007fisher, jegou2010aggregating}. 
Note that we do not impose any constraints on the sparsity of $\bf x$ in the direct domain or some transform domain. 
%The template ${\bf y} \in {\mathcal{Y}}^d$ is defined in the domain ${\mathcal{Y}}$ to be a space of real or discrete (often binary) representations with $d \leq n$. 

%The template extractor $\phi_{\boldsymbol \theta}(\cdot)$ can be any network like AlexNet \cite{krizhevsky2012imagenet}, VGG$\_$19 \cite{simonyan2014very}, etc., or a latent representation of auto-encoders. 
In this work, we will focus on a simple model $\phi_{\boldsymbol \theta}({\bf x}) = W{\bf x}$ to investigate the capacity of machine learning methods versus analytical closed-form solutions for reconstruction attacks against protected templates. Our setup is explained by the need to investigate the security of template protection rather than template extraction since the reconstruction from various features and templates is known to be successful according to \cite{dosovitskiy2016inverting, mai2018reconstruction}. 
%with known template extraction parameters. 
%However, the described reconstruction algorithms apply to all $\phi_{\boldsymbol \theta}({\cdot})$ with the constrains which are mentioned below.

%%% generic Privacy preserinv %%%%
As privacy protection, we consider a generic privacy-preserving encoding based on random projections, quantization and addition of random noise that is integrated in the STCA framework \cite{Razeghi2017wifs, Razeghi2018icassp}:\vspace{-8pt}
\begin{equation}
\label{eq:privacy_preserving_model}
{\bf u}_a = \varphi(A {\bf y}) \oplus  \mathbf{n}_{\mathrm{supp}},
\end{equation}
where  $A$ is a random $m \times d$ matrix, $\varphi(\cdot):\mathcal{R} \rightarrow \{-1,0,+1\}$ is a quantization operator representing an element-wise non-linearity and $ \mathbf{n}_{\mathrm{supp}}$ denotes the ambiguization noise that can be added to the orthogonal complement of space of $ \varphi(A {\bf y})$. The sparse ternary representation ${\bf u} = \varphi(A {\bf y})\in \{-1,0,+1\}^m$ has $S_x$ non-zero components, i.e., $\Vert {\bf u} \Vert_0 = S_x$, and protected template ${\bf u}_a$ has $S_x +  S_{n_s}$ non-zero components, where $0 \! \leq \! S_{n_s} 
\! \leq \! m \!- \! S_x$ is the sparsity level of ambiguization noise for the public representations. % \ref{eq:privacy_preserving_model}
Note that $A$ may reduce, keep or extend the dimension of the vector $\mathbf{y}$, i.e., $ m \gtreqqless d$. 
The clean representation ${\bf u}$ will be used for adversarial training.

% Recovery problem and assumptions
In this paper, our purpose is to evaluate practically the privacy-protection capacity of STCA based template protection by utilizing machine learning reconstruction tools. The general block diagram of our framework is depicted in Fig.~\ref{Fig:GeneralBlockDiagram}. 
We investigate both synthetic and real data in our study. 
We consider two attacking scenarios assuming that: (a) $\varphi(\cdot)$, $A$ and $W$ are known to the attacker and $\mathbf{n}_{\mathrm{supp}}$ is unknown and (b) $\varphi(\cdot)$, $A$, $W$, $ \mathbf{n}_{\mathrm{supp}}$ are unknown to the attacker. The goal of the attacker is to reconstruct $\bf \hat x$ as close as possible to $\bf x$ based on the protected template ${\bf u}_a$.

%$\mathbf{y} = \sigma_1 ( W_1 \mathbf{x}) $ where $ W_1$  is a random matrix and $\sigma_1( \cdot )$ is a STC element-wise non-linearity.

\vspace{-9pt}

\section{Reconstruction under the known model parameters}
\label{sec:recovery_known}

\vspace{-8pt}

In the most general case, the reconstruction under the known models (\ref{eq:template extraction}) and (\ref{eq:privacy_preserving_model}) can be formulated as:\vspace{-6pt}
\begin{equation}
\label{eq:ps_inv_reg0}
{\bf \hat x} = \arg \min_{\bf x}  \frac{1}{2}||{\bf u}_a  - \varphi(A\phi_{\boldsymbol \theta}({\bf x})) ||^2_2  + \lambda \Omega({\bf x}),
\end{equation}
where $||.||_2$ denotes a $\ell_2$-norm, $\! \lambda$ is a regularization parameter and $ \lambda \Omega({\bf x}) \! \! =\!    - \! \log p({\bf x})$. Unfortunately, it is not feasible to define a multidimensional pdf $p({\bf x})$ in practice besides some rare exceptions of i.i.d. models and the models capturing~sparsity. 
%, simple models capturing local correlations and potentially sparsity in some defined or learnable transform domain. 

To ensure privacy protection against adversarial reconstruction, we separate the impact of the imposed non-linearity by the network $\phi_{\boldsymbol \theta}(\cdot)$ on the extracted templates, from the STCA template protection capacity. Accordingly, we~consider the network $\phi_{\boldsymbol \theta}({\bf x}) \! =\! W{\bf x}$ with a constraint $W^TW \! = \! I$, that yields:\vspace{-13pt}
\begin{equation}
\label{eq:ps_inv_reg1}
{\bf \hat x} = \arg \min_{\bf x}  \frac{1}{2}||{\bf u}_a  - \varphi(AW {\bf x}) ||^2_2  + \lambda \Omega({\bf x}).
\end{equation}

In the general case, assuming that $\varphi(\cdot)$ is differentiable with respect to $\bf x$, one can find the solution to (\ref{eq:ps_inv_reg1}) as:\vspace{-6pt}
\begin{equation}
\label{eq:note_adv_example_learning}
 {\bf x}  \gets  {\bf x} - \eta \, \nabla_{\! \bf x} J({\bf x}),
 \end{equation}
by denoting $J({\bf x}) = \frac{1}{2}||{\bf u}_a  - \varphi(AW {\bf x}) ||^2_2  + \lambda \Omega({\bf x})$.
However, the solution of (\ref{eq:ps_inv_reg1})  based on (\ref{eq:note_adv_example_learning}) faces several practical problems that we summarize below.

\vspace{-2pt} 

{\em Problem 1 (the non-differentiability of $\varphi(\cdot)$)}:  The STCA ternarization operator  $\varphi(\cdot)$ is  not differentiable  with respect to $\bf x$. Therefore, one can envision several approaches to approximate $\varphi(\cdot)$ by some differentiable surrogate function, as for example by a hard-thresholding operator that preserves the same mutual information as for the ternarization operator for a range of sparsity levels $S_x$ \cite{1901.08437}, or by considering a linear approximation, i.e., $\varphi(AW {\bf x})=AW {\bf x}$, that yields to:\vspace{-7pt}
 \begin{eqnarray}
\label{eq:ps_inv_reg_linearization}
{\bf \hat x} = \arg \min_{\bf x}  \frac{1}{2}||{\bf u}_a  - AW {\bf x} ||^2_2  + \lambda \Omega({\bf x}),
\end{eqnarray}
with $\Omega({\bf x}) = || {\bf x}||_2^2$. In this case, the solution reduces to:\vspace{-5pt}
\begin{equation}
\label{eq:pseudo_inverse}
{\bf \hat x} = ((AW)^TAW + \lambda I)^{-1}(AW)^T{\bf u}_a.
\end{equation}

\vspace{-5pt} 

{\em Problem 2 (model prior for real data)}: The above assumed $\ell_2$-norm regularizer works only for the synthetic i.i.d Gaussian data. In the case of real images, it is too restrictive. The class of sparsification priors is also relatively restrictive in view of a single overcomplete shallow representation. Instead, recent works \cite{bora2017compressed,bojanowski2017optimizing} suggested using a generative model ${\bf x} = g_{\boldsymbol \theta_G} ({\bf z})$, where $g_{\boldsymbol \theta_G} (\cdot)$ is a generator of GAN or decoder of VAE trained on corresponding data $\{{\bf x}_i\}_{i=1}^N$, where $N$ denotes the number of training samples. In this case, the reconstruction problem reduces to:\vspace{-8pt}
\begin{equation}
\label{eq:ps_inv_generative}
{\bf \hat z} = \arg \min_{\bf z}  \frac{1}{2}||{\bf u}_a  - \varphi(AW g_{\boldsymbol \theta_G} ({\bf z})) ||^2_2  + \lambda \Omega({\bf z}),
\end{equation}
and ${\bf \hat x} = g_{\boldsymbol \theta_G} ({\bf \hat z})$ under the differentiability or surrogate replacement of non-differentiable $\varphi(\cdot)$.

\vspace{-8pt}

\section{Reconstruction under the unknown model parameters}
\label{sec:recovery_unknown}
\vspace{-7pt}

 The adversarial reconstruction under the model with unknown parameters is of more practical interest. Instead of assuming the complete knowledge of the model and its differentiability, the adversary has access to the training data $\{{\bf x}_j, {\bf u}_j\}_{j=1}^M$ or can use the model ${\bf u}= \varphi(A {\bf y})$ as a {\em black box} for the given inputs $\{{\bf x}_j\}_{j=1}^M$ to compute the templates $\{{\bf  u}_j\}_{j=1}^M$.  

The recovery problem reduces to the training of a reconstruction deep network $f_{{{\boldsymbol \psi}}}( {\bf u})$:\vspace{-10pt}
%According to the machine learning recovery approach, a decoder $g_{{\boldsymbol {\boldsymbol \theta}}}(.)$  is trained on the training dataset $(\mathcal{D}_x, \mathcal{D}_y  )$:
\begin{eqnarray}
\label{eq:DNN_decoder_clean}
 {\boldsymbol \psi}^*= \arg \min_{{ {\boldsymbol \psi}}} \frac{1}{M} \sum_{j=1}^M {\mathcal L}  ( f_{{{\boldsymbol \psi}}}( {\bf u}_j), {\bf x}_j  ) .
%+\lambda \Omega_{{\boldsymbol {\boldsymbol {\boldsymbol \theta}}}_D }({\boldsymbol {\boldsymbol {\boldsymbol \theta}}}_D )
\end{eqnarray}
The trained reconstruction network represents a deep decoder that is applied directly to a privacy-protected template ${\bf u}_a$ by producing the recovered image ${\bf \hat x} \! =\! \! f_{{\boldsymbol \psi}^*}({\bf  u}_a)$. One can also envision a setup of training on protected templates $\mathbf{u}_a$ with different ambiguization levels of $\mathbf{u}$ that is out of the scope of this paper. 

\vspace{-10pt}

\section{Experimental results}
\label{sec:Experimental-results}

\vspace{-5pt}

The privacy protection power of STCA originates from the lossy ternary quantization induced by $\! \varphi(\cdot)$ and the addition of the ambiguization noise $\mathbf{n}_{\mathrm{supp}}$ to the support complement of sparse ternary approximation. 
Clearly, the imposed lossy ternarization has an impact on both the authorized and unauthorized users that prevents the perfect recovery of $\bf x$. 
We can measure how much information is lost and how much is preserved, in the terms of distortion level and encoding rate. Moreover, we can link it to the classical Shannon rate-distortion theory. 
The intelligently designed ambiguization scheme proposed by the STCA ensures accurate reconstruction for the authorized users while prohibits an accurate reconstruction for the unauthorized users. This originates from the fact that the authorized users can estimate the correct support of the data using their noisy data $\mathbf{x} + \mathbf{z}$, while the unauthorized users have no knowledge to ``unlock'' the protected template.

%should prevent the recovery for the unauthorized users while preserve the information and thus ensure the accurate recovery for the authorized users. 

Consequently, we will investigate: (a) the link to the rate-distortion function, (b) the reconstruction based on the pseudo-inverse \eqref{eq:pseudo_inverse} and (c) the reconstruction based on the trained network \eqref{eq:DNN_decoder_clean}. To evaluate the achievability of theoretical limits, we will first validate our results on synthetic data and then extend them to real images. As the synthetic data, we used i.i.d. Gaussian samples ${\bf X} \sim \mathcal{N} ({\bf 0}, \sigma_X^2 I_n)$ and mappers $A_{i,j} \sim \mathcal{N} (0, 1/\sqrt{n})$ and DCT transform as $W$.  As the real images, we have used MNIST data set. Although, this dataset is relatively simple however it is very useful to assess the quality of reconstruction in terms of unique recognition of digits by human thus avoiding any subjective factors of quality evaluation. The mappers $A$ and $W$ are the same as for the synthetic data. 

\begin{figure}[t]
    \centering
     \hspace{-9pt}
        \begin{subfigure}[h]{0.24\textwidth}
        \includegraphics[width=4.64cm,height=3.3cm]{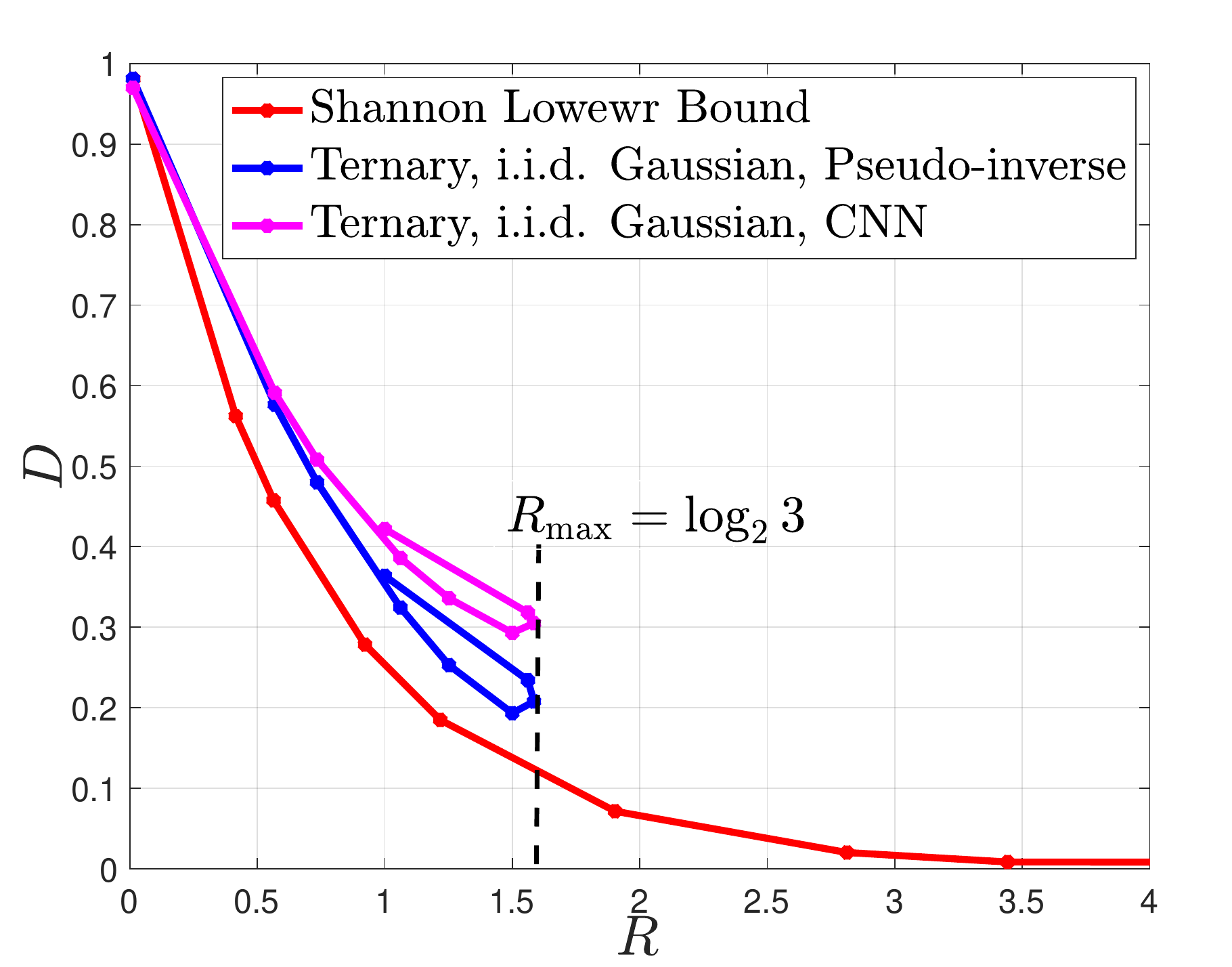}%
           \vspace{-6pt}
        \caption{}
           \vspace{-3pt}
        \label{fig:DR-iidGaussian-limit-synthetic-data}
    \end{subfigure}%
 ~%%
       \begin{subfigure}[h]{0.24\textwidth}
        \includegraphics[width=4.64cm,height=3.3cm]{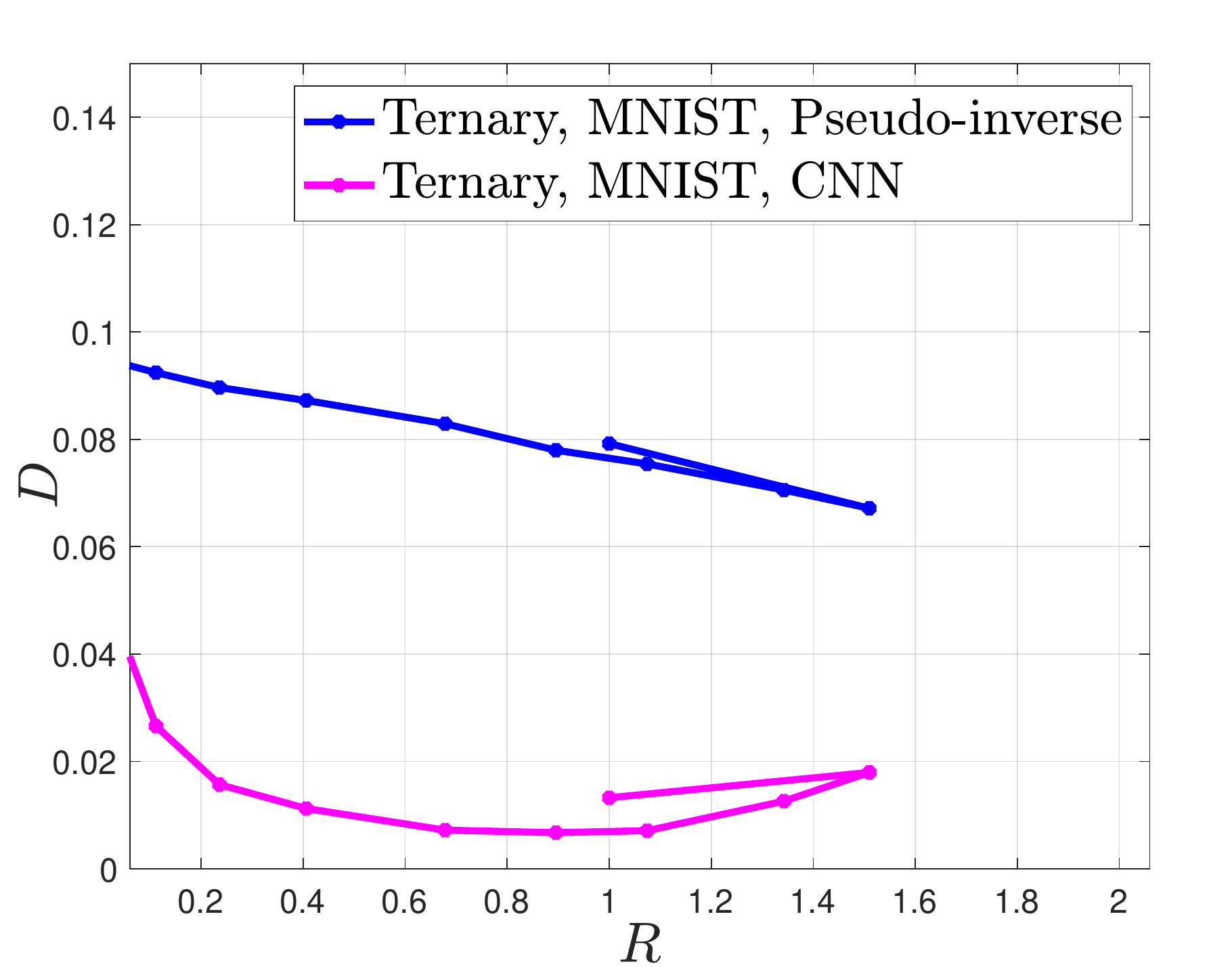}%
        \vspace{-6pt}
        \caption{}
           \vspace{-3pt}
        \label{fig:DR-iidGaussian-limit-real-data}
    \end{subfigure}
    
    \hspace{-9pt}
    \begin{subfigure}[h]{0.24\textwidth}
        \includegraphics[width=4.65cm,height=3.3cm]{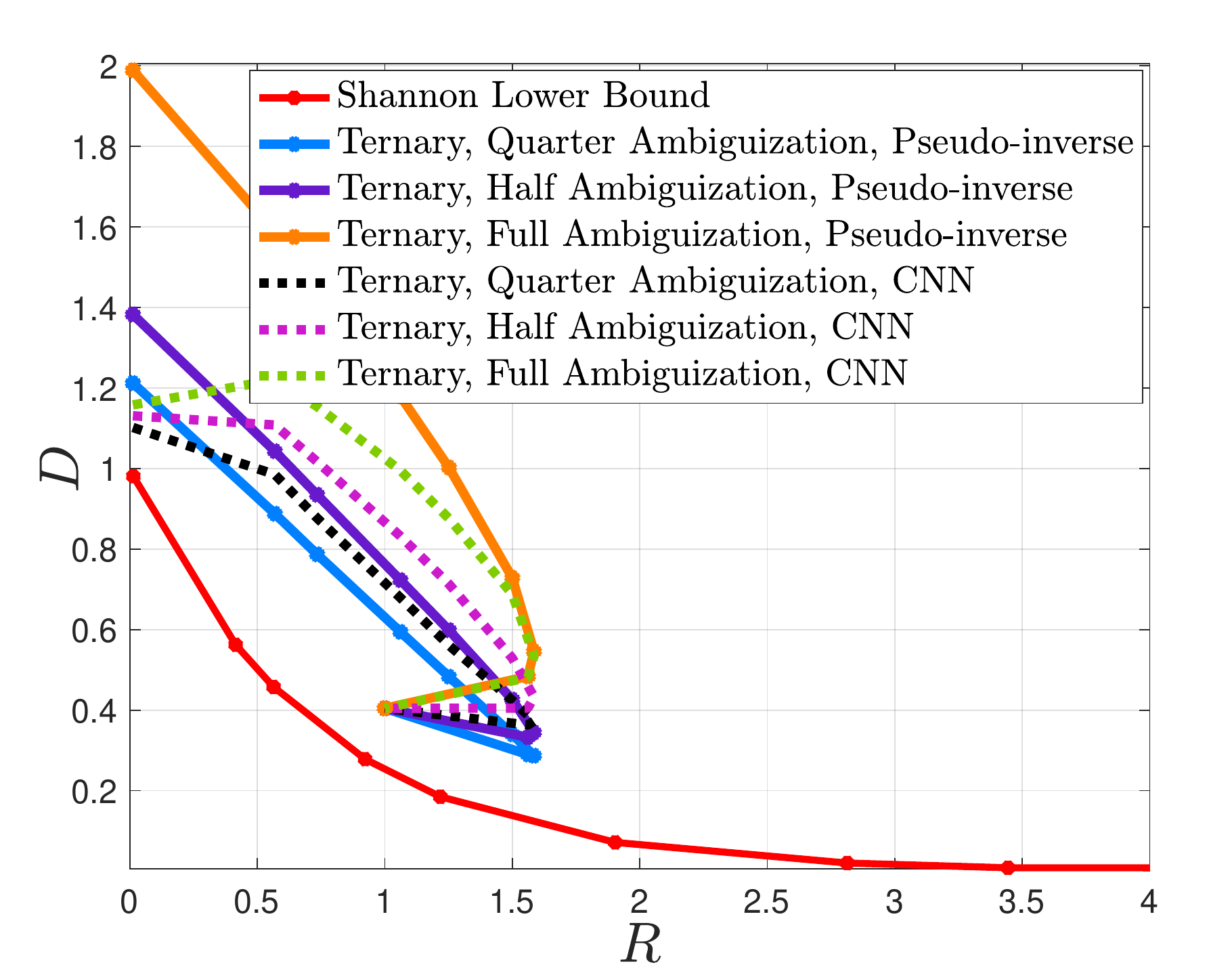}%
       \vspace{-6pt}
        \caption{}
           \vspace{-3pt}
        \label{fig:DR-unauthorized-synthetic-data}
    \end{subfigure}%
    ~%%
     \begin{subfigure}[h]{0.24\textwidth}
        \includegraphics[width=4.64cm,height=3.3cm]{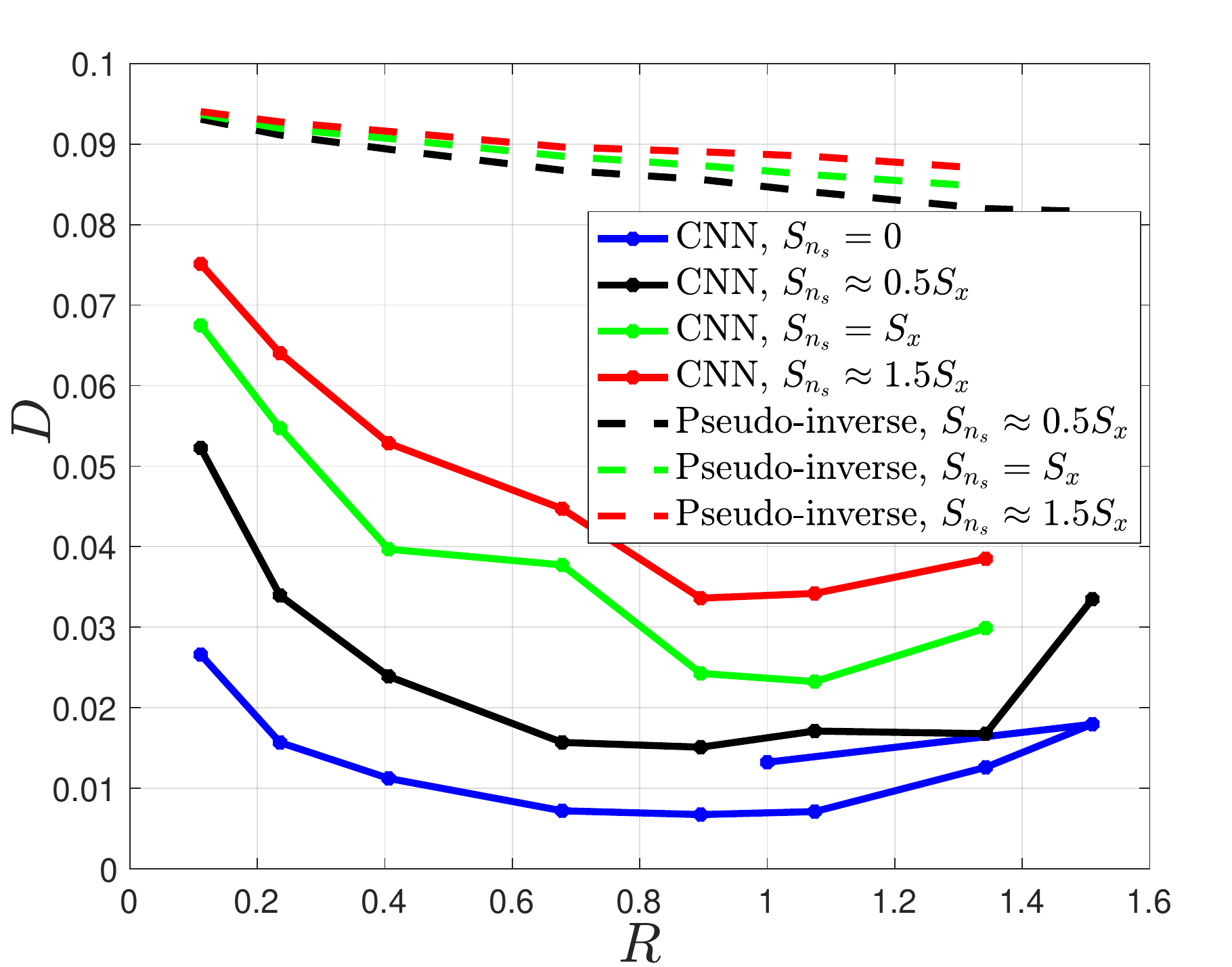}%
        \vspace{-6pt}
        \caption{}
           \vspace{-3pt}
        \label{fig:DR-unauthorized-real-data}
    \end{subfigure}
    
    \hspace{-9pt}
    \begin{subfigure}[h]{0.24\textwidth}
        \includegraphics[width=4.65cm,height=3.3cm]{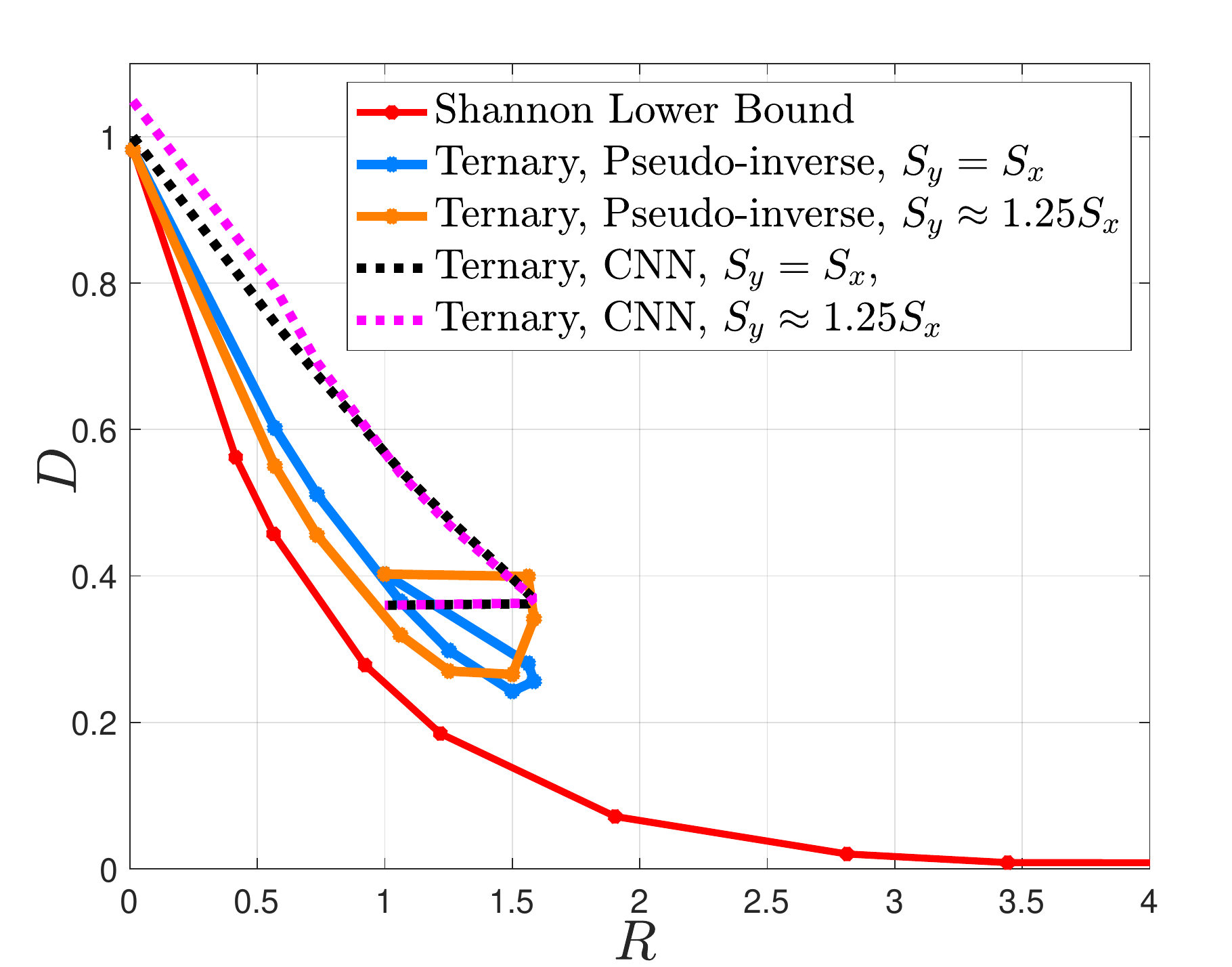}%
        \vspace{-6pt}
        \caption{}
           \vspace{-10pt}
        \label{fig:DR-Authorized-synthetic-data}
    \end{subfigure}%
    ~%%
     \begin{subfigure}[h]{0.24\textwidth}
        \includegraphics[width=4.64cm,height=3.3cm]{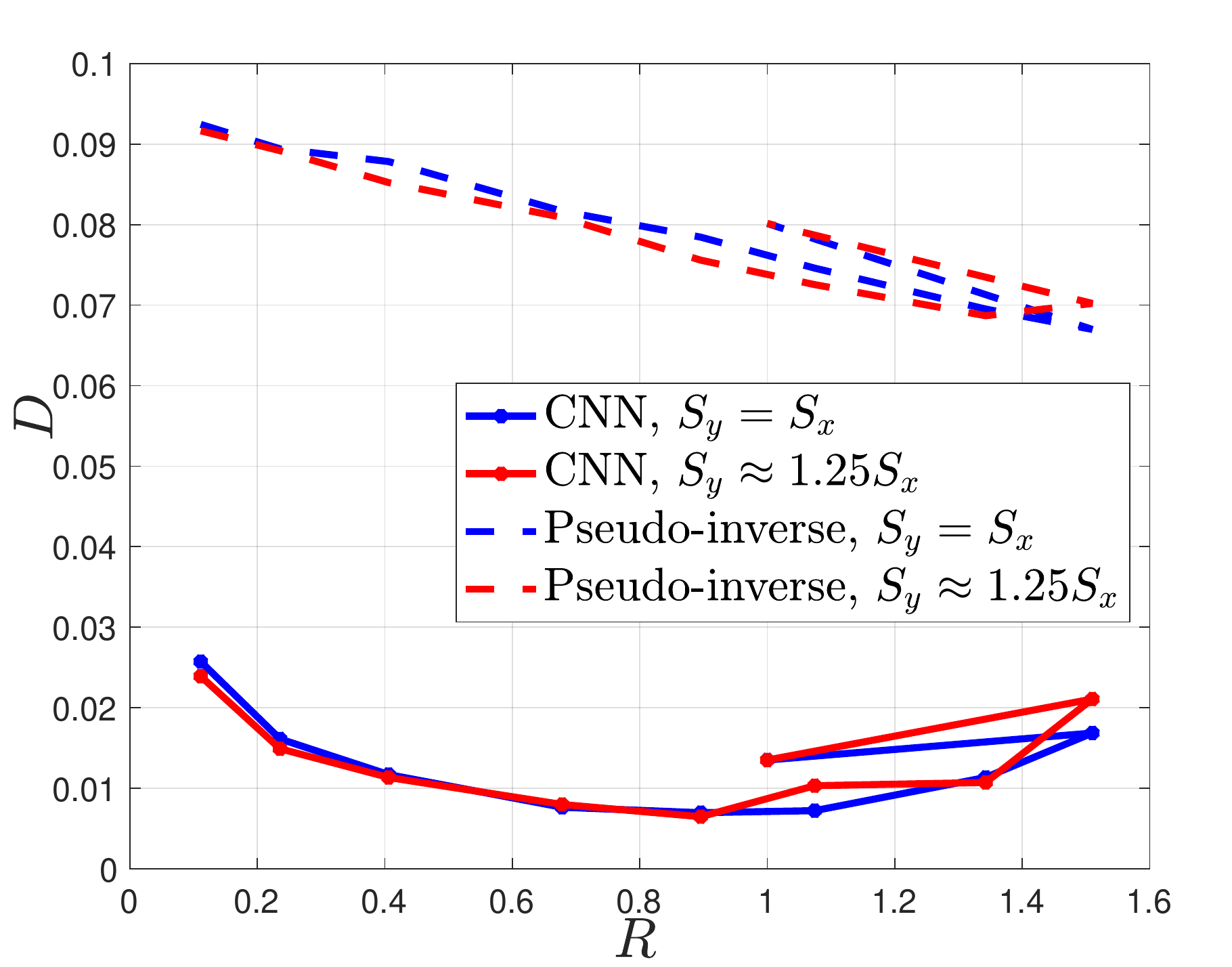}%
        \vspace{-6pt}
        \caption{}
           \vspace{-10pt}
        \label{fig:DR-Authorized-real-data}
    \end{subfigure}
    \caption{Comparison of Rate-Distortion behavior on the synthetic and real data sets. (a), (c), (e): comparison of pseudo-inverse and CNN reconstructions on synthetic data set. (b), (d), (f): comparison of pseudo-inverse and deep reconstructions on MNIST data set. (a), (b): $R(D)$ curve limits. (c), (d): $R(D)$ curve for unauthorized users with different ambiguization levels. (e), (f): $R(D)$ curve for authorized users with different sparsity levels $S_y$ and considering noisy measurement $ \mathbf{x} + \mathbf{z}$, such that $\sigma^2_Z = 0.25 \, \sigma^2_X$.}
    \vspace{-20pt}
    \label{Fig:DisPre}
\end{figure}

We present three series of tests: (i) the theoretical limits of reconstruction from the STC representations in terms of achieving the Shannon lower bound on rate-distortion \cite{cover2012elements} is investigated in Figures \ref{fig:DR-iidGaussian-limit-synthetic-data} and \ref{fig:DR-iidGaussian-limit-real-data}, (ii) the capability of adversary to reconstruct from the protected templates ${\bf u}_a$ with different levels of sparsity is depicted in Figures \ref{fig:DR-unauthorized-synthetic-data} and \ref{fig:DR-unauthorized-real-data} and (iii) the accuracy of reconstruction from the protected templates for authorized users with access to the noisy data is shown in Figures \ref{fig:DR-Authorized-synthetic-data} and \ref{fig:DR-Authorized-real-data}. 

%his information advantage over the adversary having an access to the noisy data $\bf x+z$ that allows to estimate the support and un-lock the data with certain accuracy shown in Figures \ref{Fig:synthetic_data}c and \ref{Fig:real_data}c. 

All results are shown for the reconstruction based on the pseudo-inverse (\ref{eq:pseudo_inverse}) with $\lambda \! \! =\! \! 0$ and two types of deep reconstruction networks. The reconstruction network (\ref{eq:DNN_decoder_clean}) for the synthetic data consists of: Linear(529) $\! \to \!$ Tanh $\to$ Reshape2D $\to \!$ Conv2D (channels=32, kernel=7) $ \! \to \!$ ReLu $ \!\to \!$ Conv2D (channels=16, kernel=5) $ \!\to \!$ Tanh $ \!\to \!$ Conv2D (channels=8, kernel=3) $ \!\to \!$ ReLu $ \!\to \!$ Conv2D (channels=1, kernel=3) $ \!\to \!$ Tanh. The reconstruction network for the real data consists of: Linear(784) $\! \to \!$ ReLu $\! \to \!$ Reshape2D $ \!\to \!$ Conv2D (channels=32, kernel=5) $ \! \to \!$ ReLu $ \! \to \!$ Conv2D(channels=16, kernel=5) $\! \to \!$ ReLu $\! \to \!$ Conv2D (channels=1, kernel=5) $\! \to \!$ ReLu.

% \begin{figure}
%     \centering
%     \includegraphics[scale=0.34]{Figures/Fig1_a.pdf}
%     \includegraphics[scale=0.34]{Figures/Fig1_b.pdf}
%     \includegraphics[scale=0.34]{Figures/Fig1_c.pdf}
%     \caption{Comparison of pseudo-inverse and deep reconstructions on MNIST images: sparsity 50 and ambiguization 25 and 100.}
%     \label{Fig:synthetic_data}
% \end{figure}

% \begin{figure}
%     \centering
%     \includegraphics[scale=0.34]{Figures/Fig1_a.pdf}
%     \includegraphics[scale=0.34]{Figures/Fig1_b.pdf}
%     \includegraphics[scale=0.34]{Figures/Fig1_c.pdf}
%     \caption{Comparison of pseudo-inverse and deep reconstructions on MNIST images: sparsity 50 and ambiguization 25 and 100.}
%     \label{Fig:real_data}
% \end{figure}

\begin{figure}
        \centering
     \hspace{-12pt}
        \begin{subfigure}[h]{0.24\textwidth}
        \includegraphics[scale=0.21]{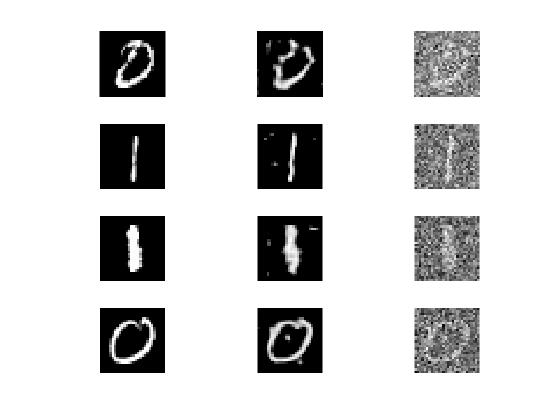}%
           \vspace{-12pt}
        \caption{}
           %\vspace{-2pt}
        \label{fig:3a-MNIST-comparison}
    \end{subfigure}%
 ~~~~~%%
       \begin{subfigure}[h]{0.24\textwidth}
        \includegraphics[scale=0.21]{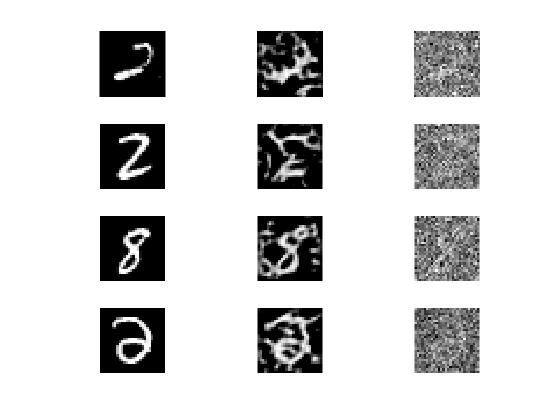}%
        \vspace{-12pt}
        \caption{}
           %\vspace{-2pt}
        \label{fig:3b-MNIST-comparison}
    \end{subfigure}
        \vspace{-12pt}
    \caption{Comparison of deep and pseudo-inverse reconstructions on MNIST images. (a): $S_x \! = \!50$, $S_{n_s} \! =  \! 25$, (b): $S_x \! = \! 50$, $S_{n_s}\! \!= \! 100$. First column of each panel corresponds to the original data, second and third ones correspond to reconstructed data based on deep network and pseudo-inverse, respectively.}
    \vspace{-11pt}
    \label{Fig:MNIST-comparison}
\end{figure}

The obtained results demonstrate that the pseudo-inverse reconstruction and deep net reconstruction for the i.i.d. synthetic data have very similar performance while for the real data the deep reconstruction benefits from the presence of structured training data and produces perceptually more pleasant results as shown in Figure \ref{Fig:MNIST-comparison}. The pseudo-inverse reconstruction does not use data priors and the produced results are quite noisy. 
%However, one can apply the post-processing and envision alternative algorithms that should be additionally investigated in future. 

It is interesting to note that both types of reconstruction strategies approach the Shannon limit on the synthetic data for the rates less than $\log_2 3 = 1.585$, which is the maximum achievable rate by the ternary encoding used in the STCA. This behaviour is depicted in Fig.~\ref{fig:DR-iidGaussian-limit-synthetic-data}. However, the pseudo-inverse reconstruction produces the results closer to the Shannon lower bound. This can be explained that the encoding matrix is well defined and known while in the case of CNN reconstruction the inverse mapping is learned from the training i.i.d data only. 
%The maximum achiveable rate by ternary encoding is $\log_2 3 = 1.585$. 

The ambiguization has a strong influence on the adversary reconstruction ability according to Fig.~\ref{fig:DR-unauthorized-synthetic-data} and Fig.~\ref{fig:DR-unauthorized-real-data}  that is clearly reflected by the increase of reconstruction distortion with the increase of the ambiguiszation $S_{n_s}$. However, in the case of real data, the structure in the data makes it possible for the unauthorized user to reconstruct from the protected template as shown in Fig.~\ref{fig:DR-unauthorized-real-data}. Several examples of the comparison between the pseudo-inverse reconstruction and those based on the trained network are shown in Fig.~\ref{Fig:MNIST-comparison}. As expected, increasing the ambiguization makes the recognition more difficult and sometimes ambiguous (Fig.~\ref{fig:3b-MNIST-comparison}). In practice, one should find the correct ambiguization factor to a particular type of data.
Finally, the authorized user can unlock the ambiguization and approach the theoretical limit in contrast to the adversary (Fig.~\ref{fig:DR-unauthorized-synthetic-data}, \ref{fig:DR-unauthorized-real-data}) as shown in Figures \ref{fig:DR-Authorized-synthetic-data} and \ref{fig:DR-Authorized-real-data}. This confirms the main research hypothesis in the paper.

\vspace{-6pt}

\section{Conclusions}
\label{Conclusions}

\vspace{-8pt}

We studied the capability of STCA for privacy-protection data release. 
The validates shown that for synthetic data, the STCA provides the theoretically achievable limits under both known and unknown model based adversarial reconstruction. 
Moreover, neglection of the ternarization non-linearity and usage of a simple pseudo-inverse does not lead to any drop in performance in comparison to deep non-linear reconstruction.
An interesting point is that the deep reconstruction algorithm does not benefit from i.i.d data in comparison to simple projection matrix knowledge used in the pseudo-inverse. 

The analysis on the real images established that the difference between the accuracy of reconstruction produced by the linear pseudo-inverse and deep reconstruction is significant besides the fact that the pseudo-inverse uses priors about the models of template extraction and template protection and deep reconstruction is based solely on the training data. 
We consider two major factors that impact this difference: (i) the fact that the linear pseudo-inverse does not use any prior about the statistics of images while the deep reconstruction can learn data manifold implicitly from the training data, (ii) the non-linear deep reconstruction also overcomes the problem of differentiability of the ternarization operator while the linear pseudo-inverse used in this paper uses the linear approximation. 
We believe that these factors should be covered in future research. In particular, the lack of reliably image priors can be addressed by the generative model in the formulation (\ref{eq:ps_inv_generative}) under the linear ternarization approximation and compared with the solution based on (\ref{eq:DNN_decoder_clean}). 
An additional research problem to be addressed is an \textit{adversarial training} (\ref{eq:DNN_decoder_clean}), when the various ambiguizations ${\bf u}_a$ should replace the clean version ${\bf u}$ similarly to the denoising auto-encoder training \cite{Vincent2010StackedDA}.

%%%%%%%%%%%%%%%%%%%%%%%%%%%%%%%%%%%%%

%
%\section{COPYRIGHT FORMS}
%\label{sec:copyright}
%
%You must include your fully completed, signed IEEE copyright release form when
%form when you submit your paper. We {\bf must} have this form before your paper
%can be published in the proceedings.
%
%\section{REFERENCES}
%\label{sec:ref}
%
%List and number all bibliographical references at the end of the
%paper. The references can be numbered in alphabetic order or in
%order of appearance in the document. When referring to them in
%the text, type the corresponding reference number in square
%brackets as shown at the end of this sentence \cite{C2}. An
%additional final page (the fifth page, in most cases) is
%allowed, but must contain only references to the prior
%literature.
%
%% References should be produced using the bibtex program from suitable
%% BiBTeX files (here: strings, refs, manuals). The IEEEbib.bst bibliography
%% style file from IEEE produces unsorted bibliography list.
% -------------------------------------------------------------------------

\clearpage
\newpage
%\pagebreak

\bibliographystyle{IEEEbib}
\bibliography{refs}

\begin{thebibliography}{10}

\bibitem{dosovitskiy2016inverting}
Alexey Dosovitskiy and Thomas Brox,
\newblock ``Inverting visual representations with convolutional networks,''
\newblock in {\em Proceedings of the IEEE Conference on Computer Vision and
  Pattern Recognition}, 2016, pp. 4829--4837.

\bibitem{mai2018reconstruction}
Guangcan Mai, Kai Cao, C~YUEN Pong, and Anil~K Jain,
\newblock ``On the reconstruction of face images from deep face templates,''
\newblock {\em IEEE Transactions on Pattern Analysis and Machine Intelligence},
  2018.

\bibitem{dodis2004fuzzy}
Yevgeniy Dodis, Leonid Reyzin, and Adam Smith,
\newblock ``Fuzzy extractors: How to generate strong keys from biometrics and
  other noisy data,''
\newblock in {\em International conference on the theory and applications of
  cryptographic techniques}. Springer, 2004, pp. 523--540.

\bibitem{bringer2008theoretical}
Julien Bringer, Herv{\'e} Chabanne, Gerard Cohen, Bruno Kindarji, and Gilles
  Zemor,
\newblock ``Theoretical and practical boundaries of binary secure sketches,''
\newblock {\em IEEE Transactions on Information Forensics and Security}, vol.
  3, no. 4, pp. 673--683, 2008.

\bibitem{verbitskiy2003reliable}
Evgeny Verbitskiy, Pim Tuyls, Dee Denteneer, and Jean-Paul Linnartz,
\newblock ``Reliable biometric authentication with privacy protection,''
\newblock in {\em Proc. 24th Benelux Symposium on Information theory}, 2003,
  p.~19.

\bibitem{ignatenko2009biometric}
Tanya Ignatenko and Frans~MJ Willems,
\newblock ``Biometric systems: Privacy and secrecy aspects,''
\newblock {\em IEEE Transactions on Information Forensics and security}, vol.
  4, no. 4, pp. 956, 2009.

\bibitem{sutcu2005secure}
Yagiz Sutcu, Husrev~Taha Sencar, and Nasir Memon,
\newblock ``A secure biometric authentication scheme based on robust hashing,''
\newblock in {\em Proceedings of the 7th workshop on Multimedia and security}.
  ACM, 2005, pp. 111--116.

\bibitem{nandakumar2015biometric}
Karthik Nandakumar and Anil~K Jain,
\newblock ``Biometric template protection: Bridging the performance gap between
  theory and practice,''
\newblock {\em IEEE Signal Processing Magazine}, vol. 32, no. 5, pp. 88--100,
  2015.

\bibitem{Razeghi2017wifs}
Behrooz Razeghi, Slava Voloshynovskiy, Dimche Kostadinov, and Olga Taran,
\newblock ``Privacy preserving identification using sparse approximation with
  ambiguization,''
\newblock in {\em IEEE International Workshop on Information Forensics and
  Security (WIFS)}, Rennes, France, December 2017, pp. 1--6.

\bibitem{Razeghi2018icassp}
Behrooz Razeghi and Slava Voloshynovskiy,
\newblock ``Privacy-preserving outsourced media search using secure sparse
  ternary codes,''
\newblock in {\em IEEE International Conference on Acoustics, Speech and Signal
  Processing (ICASSP)}, Calgary, Alberta, Canada, April 2018.

\bibitem{Razeghi2018eusipco}
Behrooz Razeghi, Slava Voloshynovskiy, Sohrab Ferdowsi, and Dimche Kostadinov,
\newblock ``Privacy-preserving identification via layered sparse code design:
  Distributed servers and multiple access authorization,''
\newblock in {\em 26th European Signal Processing Conference (EUSIPCO)}, Rome,
  Italy, September 2018.

\bibitem{Ferd1706:Sparse}
Sohrab Ferdowsi, Sviatoslav Voloshynovskiy, Dimche Kostadinov, and Taras
  Holotyak,
\newblock ``Sparse ternary codes for similarity search have higher coding gain
  than dense binary codes,''
\newblock in {\em IEEE International Symposium on Information Theory (ISIT)},
  Aachen, Germany, 2017.

\bibitem{simonyan2014very}
Karen Simonyan and Andrew Zisserman,
\newblock ``Very deep convolutional networks for large-scale image
  recognition,''
\newblock {\em arXiv preprint arXiv:1409.1556}, 2014.

\bibitem{krizhevsky2012imagenet}
Alex Krizhevsky, Ilya Sutskever, and Geoffrey~E Hinton,
\newblock ``Imagenet classification with deep convolutional neural networks,''
\newblock in {\em Advances in neural information processing systems}, 2012, pp.
  1097--1105.

\bibitem{jegou2009burstiness}
Herv{\'e} J{\'e}gou, Matthijs Douze, and Cordelia Schmid,
\newblock ``On the burstiness of visual elements,''
\newblock in {\em IEEE Conf. on Comp. Vision and Pattern Recog. (CVPR)}, 2009,
  pp. 1169--1176.

\bibitem{perronnin2007fisher}
Florent Perronnin and Christopher Dance,
\newblock ``Fisher kernels on visual vocabularies for image categorization,''
\newblock in {\em IEEE Conf. on Comp. Vision and Pattern Recog. (CVPR)}, 2007,
  pp. 1--8.

\bibitem{jegou2010aggregating}
Herv{\'e} J{\'e}gou, Matthijs Douze, Cordelia Schmid, and Patrick P{\'e}rez,
\newblock ``Aggregating local descriptors into a compact image
  representation,''
\newblock in {\em IEEE Conf. on Comp. Vision and Pattern Recog. (CVPR)}, 2010.

\bibitem{1901.08437}
Sohrab Ferdowsi,
\newblock ``Learning to compress and search visual data in large-scale
  systems,'' 2019.

\bibitem{bora2017compressed}
Ashish Bora, Ajil Jalal, Eric Price, and Alexandros~G Dimakis,
\newblock ``Compressed sensing using generative models,''
\newblock {\em arXiv preprint arXiv:1703.03208}, 2017.

\bibitem{bojanowski2017optimizing}
Piotr Bojanowski, Armand Joulin, David Lopez-Paz, and Arthur Szlam,
\newblock ``Optimizing the latent space of generative networks,''
\newblock {\em arXiv preprint arXiv:1707.05776}, 2017.

\bibitem{cover2012elements}
Thomas~M Cover and Joy~A Thomas,
\newblock {\em Elements of information theory},
\newblock John Wiley \& Sons, 2012.

\bibitem{Vincent2010StackedDA}
Pascal Vincent, Hugo Larochelle, Isabelle Lajoie, Yoshua Bengio, and
  Pierre-Antoine Manzagol,
\newblock ``Stacked denoising autoencoders: Learning useful representations in
  a deep network with a local denoising criterion,''
\newblock {\em Journal of Machine Learning Research}, vol. 11, pp. 3371--3408,
  2010.

\end{thebibliography}

\end{document}